\documentclass{style}
\usepackage{amsmath,amssymb}
\usepackage{environ}
\usepackage{fontawesome5}
\usepackage{tabularx}
\renewcommand{\emph}[1]{\textit{#1}}

\fancypagestyle{firststyle}{%
  \fancyhf{}%
  \fancyfoot[C]{\thepage}%
}
\hypersetup{
  pdftitle={PhysLoop: Harnessing LLMs' Physical Awareness for Self-Improving Video Generation Models},
  pdfauthor={Yuxuan Chen and Xiaojie Li and Hao Chen and Juncheng Ma and Kewei Zhang and Zhongjing Du and Guoqiang Wei and Fangfang Wang and Zhen Dong and Zhiyang Dou and Liming Zhao and Yan Zeng and Jianwei Zhang and Daquan Zhou}
}

\DeclareFontShape{T1}{sans}{b}{n}{<-> s*[1] resources/sans}{}
\DeclareFontShape{T1}{sans}{bx}{n}{<-> s*[1] resources/sans}{}
\renewcommand\authorformat[2][]{\authorfont{\sffamily\mbox{#2$^{#1}$}}}

\newcommand{\PhysLoopLogo}[2]{%
  \raisebox{-0.5\height}{\includegraphics[width=#1,height=9.5mm,keepaspectratio]{resources/logos/#2}}%
}
\newcommand{\PhysLoopAffiliationLogos}{%
  \vspace*{-7mm}%
  \noindent\hbox to \textwidth{%
    \PhysLoopLogo{29mm}{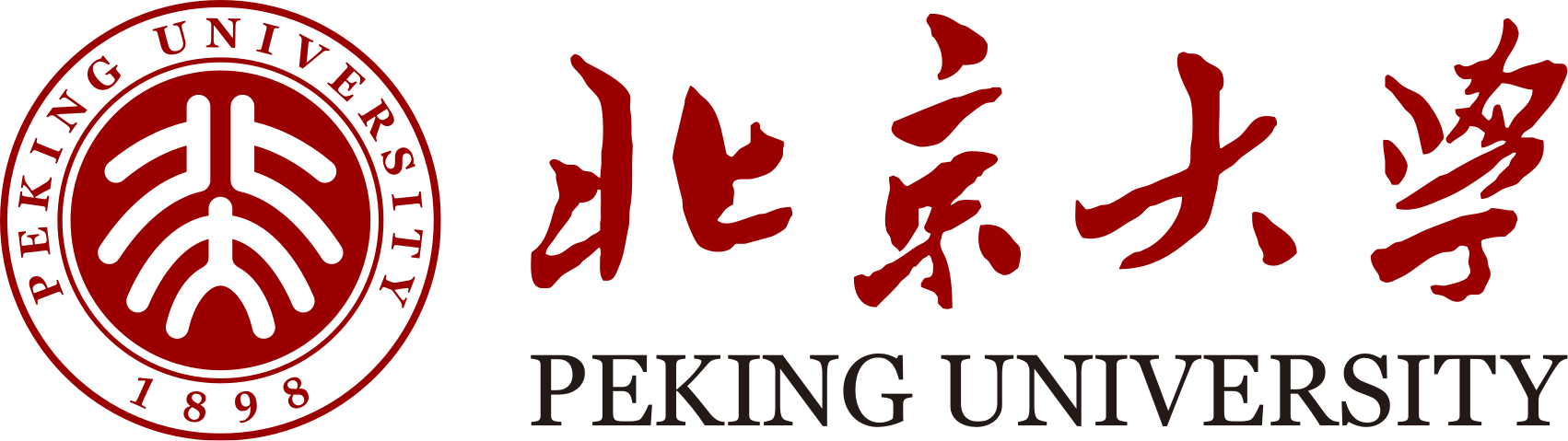}
    \raisebox{-0.5\height}{\includegraphics[width=35mm,trim=42bp 16bp 132bp 30bp,clip]{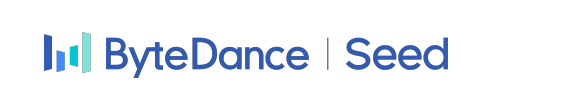}}
    \PhysLoopLogo{10mm}{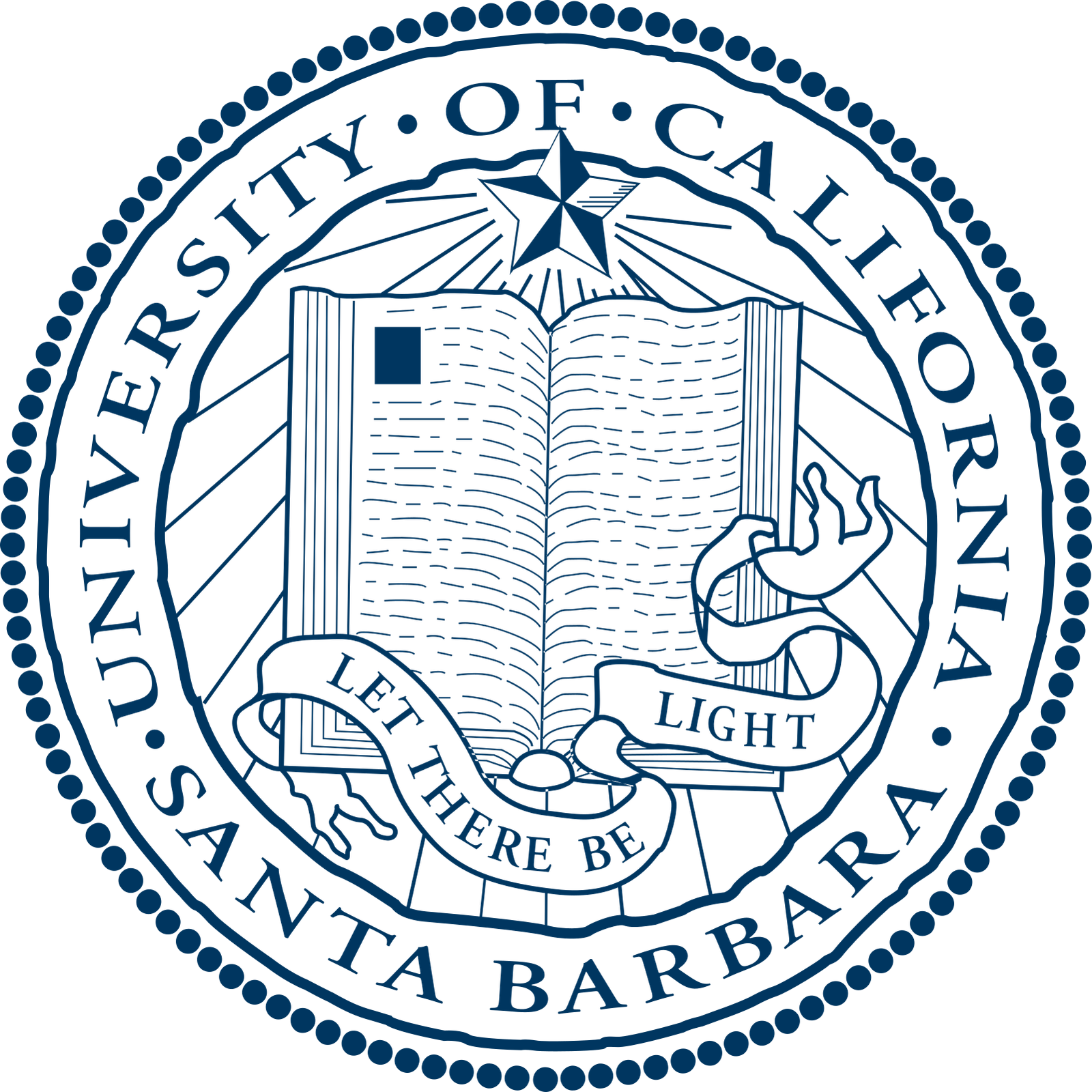}
    \PhysLoopLogo{25mm}{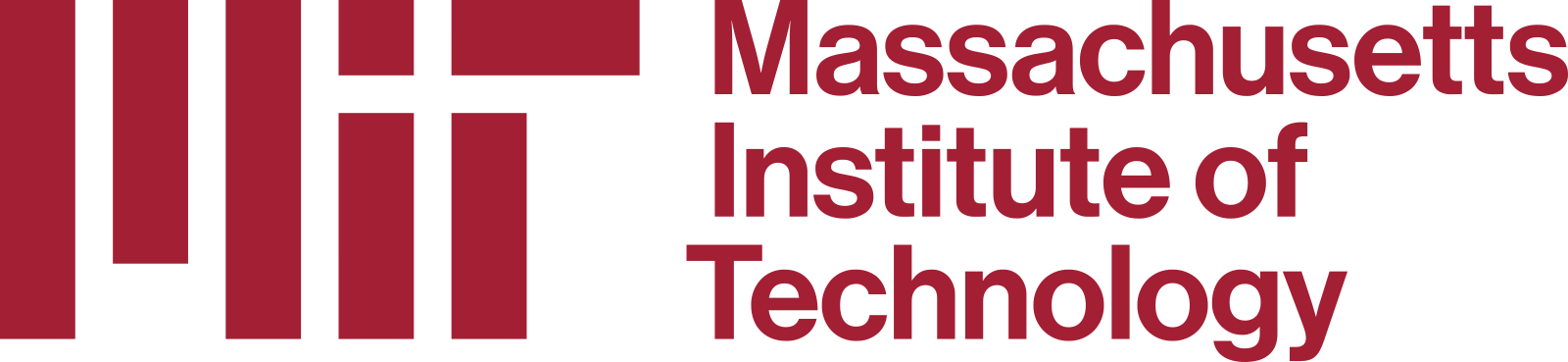}
  }%
}
\pretocmd{\mymaketitle}{\PhysLoopAffiliationLogos\par\vspace{4mm}}{}{}

\makeatletter
\renewcommand{\@toptitlebar}{\pkured{\hrule height 2.5pt}\vskip 4mm}
\renewcommand{\@bottomtitlebar}{\vskip 4mm\pkured{\hrule height 0.5pt}}
\makeatother
\patchcmd{\mymaketitle}{\vskip 6mm}{\vskip 4mm}{}{}
\patchcmd{\mymaketitle}{\vskip 6mm}{\vskip 4mm}{}{}
\patchcmd{\mymaketitle}{\vskip 3mm}{\vskip 2mm}{}{}
\patchcmd{\mymaketitle}{\vskip 3mm}{\vskip 2mm}{}{}

\DeclareRobustCommand{\PhysLoopMetaIcon}[1]{%
  \makebox[4.5mm][c]{{\fontsize{11}{11}\selectfont #1}}\hspace{1mm}%
}
\DeclareRobustCommand{\PhysLoopWebIcon}{{\color[HTML]{369CB7}\faGlobe}}
\renewcommand\checkdataformat[2][]{%
  {\fontsize{9}{13}\selectfont\strut{\sffamily\bfseries #1:}\enspace #2\strut}%
}

\title{PhysLoop: Harnessing LLMs' Physics Awareness for Self-Improving Video Generation Models}

\author[1*]{Yuxuan Chen}
\author[2*]{Xiaojie Li}
\author[1]{Hao Chen}
\author[1,2]{Juncheng Ma}
\author[1]{Kewei Zhang}
\author[1]{Zhongjing Du}
\author[2]{Guoqiang Wei}
\author[6]{Fangfang Wang}
\author[3]{Zhen Dong}
\author[4]{Zhiyang Dou}
\author[2\ddagger]{Liming Zhao}
\author[2]{Yan Zeng}
\author[5\dagger\S]{Jianwei Zhang}
\author[1\dagger\S]{Daquan Zhou}

\affiliation[1]{Peking University}
\affiliation[2]{ByteDance Seed}
\affiliation[3]{University of California, Santa Barbara}
\affiliation[4]{Massachusetts Institute of Technology}
\affiliation[5]{Universit{\normalfont\"a}t Hamburg}
\affiliation[6]{Hangzhou Normal University}

\contribution[*]{Equal contributions}
\contribution[\ddagger]{Industry advisor}
\contribution[\dagger]{Project lead}
\contribution[\S]{Corresponding authors}

\let\PhysLoopSetAbstract\abstract
\RenewEnviron{abstract}{\global\let\PhysLoopAbstractBody\BODY}
\begin{abstract}

Video generation models produce visually realistic videos but still struggle with physical dynamics and causal consistency. Learning from generated videos requires fine-grained physical supervision, yet reliable quality judgments, correction pairs, and controlled variations are difficult to obtain. Large language models (LLMs) possess physical knowledge, but struggle to reliably guide video generation models in producing their training videos. We present \textbf{PhysLoop}, a unified harness designed to support physically grounded self-improvement in video generation models. The harness guides and constrains LLMs throughout video generation, selection, and correction, distilling their physical knowledge into physically consistent videos. The framework supports controllable training-data construction and inference-time video repair. By identifying and correcting failures, PhysLoop marks a key milestone toward recursive self-improvement (RSI) in video generation models with limited human intervention. We release 100K videos spanning diverse, realistic indoor and outdoor scenes, along with generation tools and two evaluation datasets: PhysLoop-Gen and PhysLoop-Repair. We evaluate controllable generation and localized repair on these datasets, respectively, and assess physical behavior on the external Physics-IQ Verified benchmark.

\end{abstract}

\PhysLoopSetAbstract{\PhysLoopAbstractBody}

\checkdata[\protect\PhysLoopMetaIcon{\protect\PhysLoopWebIcon}Project Page]{\url{https://chenyx24.github.io/PhysLoop-site/}}
\checkdata[\protect\PhysLoopMetaIcon{\protect\faGithub}GitHub Repo]{\url{https://github.com/DAGroup-PKU/PhysLoop}}
\checkdata[\protect\PhysLoopMetaIcon{\protect\faEnvelope[regular]}Corresponding Authors]
{\protect\href{mailto:jianwei.zhang@uni-hamburg.de}{{\fontfamily{sansmedium}\selectfont Jianwei Zhang}}; \protect\href{mailto:zhoudaquan21@gmail.com}{{\fontfamily{sansmedium}\selectfont Daquan Zhou}}}

\begin{document}
\maketitle

\begin{figure}[!htbp]
  \centering
  \includegraphics[width=\linewidth]{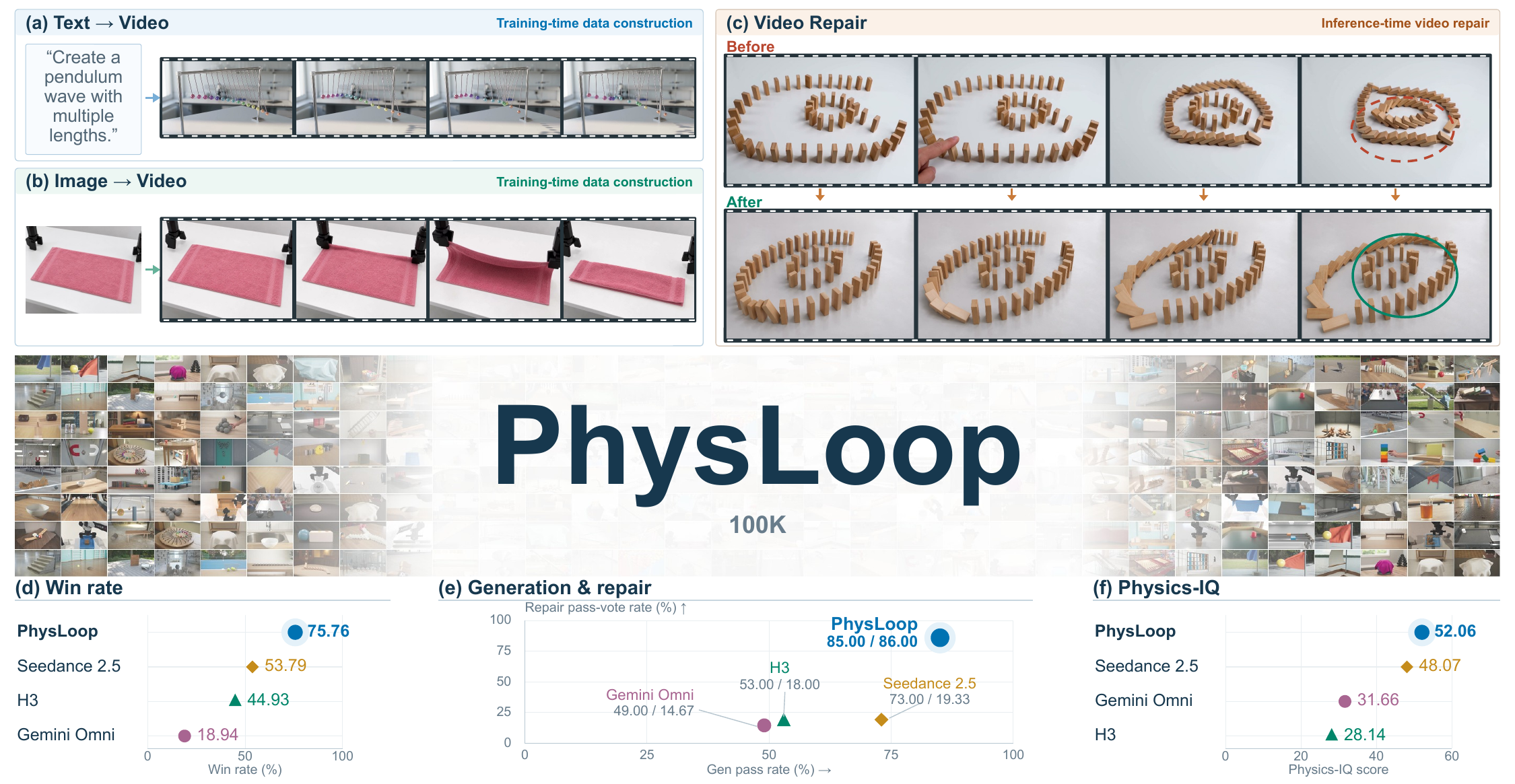}
  \label{fig:physloop-teaser}
\end{figure}

\section{Introduction}

Video generation models can produce realistic dynamic scenes with temporal coherence and viewpoint consistency~\citep{polyak2024moviegen,kong2024hunyuanvideo}. However, visual realism does not ensure physical or causal consistency: objects may disappear or multiply, and contact may fail to produce a plausible response~\citep{bansal2024videophy,bansal2026videophy2,meng2025phygenbench,motamed2026physicsiq}. Other common failures include unintended slow-motion effects and changes in speed or direction that are inconsistent with gravity, inertia, or contact forces~\citep{gu2026phyworldbench}. Such errors disrupt the temporal and causal relationships within a scene, so a video can appear convincing frame by frame while depicting a physically implausible sequence of events.

Improving physical behavior requires training videos that systematically cover how events unfold under different conditions. Such data should capture physical interactions across diverse objects and scenes~\citep{zhou2026physinone}. Collecting these videos through real-world experiments is difficult to scale while maintaining controlled conditions and reliable annotations~\citep{greff2022kubric}. Consider a glass falling and shattering on impact: varying its release height, initial velocity, or surroundings requires repeated setup, while unintended changes between trials make the effect of any one factor difficult to isolate. This collection burden also makes it difficult to obtain targeted fine-tuning data for physical behaviors that a model handles poorly, since suitable examples require deliberate coverage of the relevant conditions and reliable annotations. The challenge is therefore to obtain diverse physical videos together with dependable signals for identifying and correcting errors.

A long-term direction is recursive self-improvement (RSI), in which video generation models produce and learn from their own training data with minimal human intervention. Recent work such as KBAlign~\citep{zeng2025kbalign} and SEAL~\citep{zweiger2025seal} adapts language models using self-generated training data, while Absolute Zero~\citep{zhao2025absolutezero} learns to propose and solve tasks through verifiable execution feedback. For video generation, such a loop would first autonomously identify weaknesses in the model's physical behavior and generate candidate videos targeting them, then assess and refine the candidates and select suitable data for fine-tuning. The updated model would then generate the next round of candidates, whose quality would again be assessed. However, current video models still produce physical errors, and general-purpose multimodal evaluators can struggle to judge their physical validity reliably~\citep{bansal2026videophy2}. Generating more videos therefore does not by itself ensure that the model receives better training data. Repeated training on unverified generated data can even degrade a model across generations \cite{shumailov2024collapse}, so such a loop is only as reliable as its error detection and correction. A key requirement is to identify what is physically wrong, guide its correction, and turn the resulting videos into dependable learning signals. PhysLoop addresses this need through a unified harness that guides LLMs to generate video data in a sandbox, applies physical constraints through an internal simulator, and iteratively refines the data through physical assessment and correction.


LLMs already possess physical knowledge that supports reasoning about physical processes, but they do not apply it consistently across tasks~\citep{qiu2025phybench}. Realizing that knowledge in a video requires spatial relationships, physical conditions, and motion to remain consistent throughout an event; choices that appear plausible individually may still produce an incorrect outcome when combined~\citep{li2026physagent}. Our insight is to guide and constrain the application of this existing knowledge within an executable physical environment. The simulator inside PhysLoop supplies physical constraints and observable outcomes, allowing the harness to check whether the generated behavior satisfies the intended physical requirements. Feedback from these outcomes guides the LLM in correcting how its knowledge is applied, while repeated assessment determines which videos are suitable to retain. The accepted videos serve as physical supervision from which video generation models can learn, providing a path to distilling the LLM's physical knowledge, made executable and checked in simulation, into their generated behavior. This paper builds and evaluates this supervision; fine-tuning on it to close the loop is the next step. The same workflow supports controllable training-data construction and the assessment and repair of physical errors in generated videos.
\begin{figure*}[t]
\centering
\includegraphics[width=\linewidth]{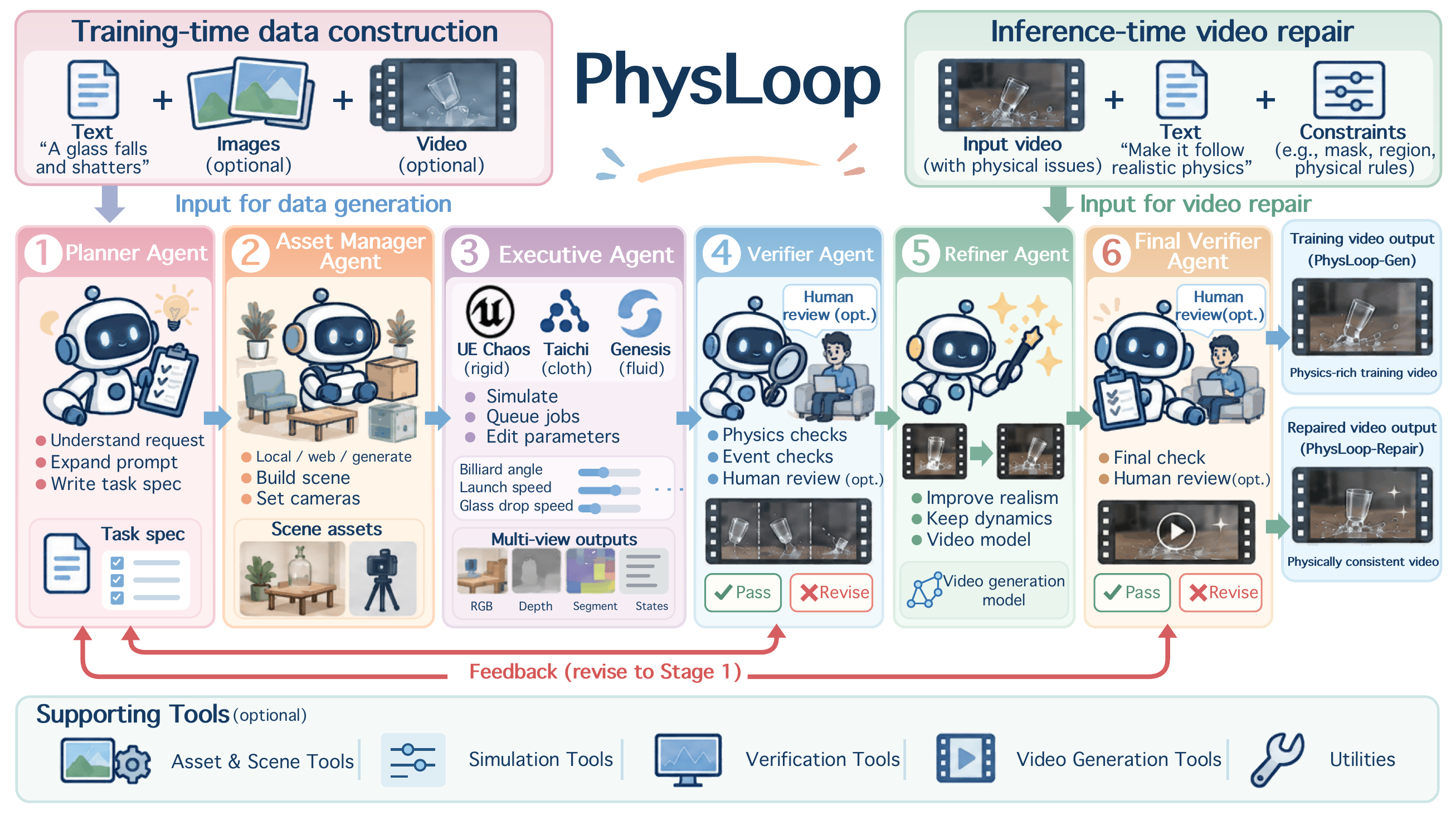}
\caption{Overview of PhysLoop and its two applications: training-data construction and video repair. The harness guides LLMs to identify physical errors and generate physically consistent videos through simulators and other tools.}
\label{fig:physloop-workflow}
\end{figure*}

We make the following contributions:
\begin{itemize}
  \item \textbf{A simulator-grounded harness toward RSI.} PhysLoop guides an LLM agent to turn its physical knowledge into executable simulations, verifies both the simulated event and the refined video against one task specification, and returns failures to the responsible stage, providing the data-construction, assessment, and repair steps that a self-improvement loop for video models requires.
  \item \textbf{Two applications of one workflow.} The harness constructs training data through controlled parameter variation and repairs physical errors in existing videos through interval replacement.
  \item \textbf{Data and benchmarks.} We build PhysLoop-100K (100K refined video--text pairs from 2.5K events, with their generating configurations and simulator depth, instance segmentation, and object-state annotations) and two evaluation sets, PhysLoop-Gen (100 tasks) and PhysLoop-Repair (50 tasks).
  \item \textbf{Evaluation.} We compare PhysLoop with three standalone video generators and five variants that replace the agent backbone or replace or remove the refiner, using blinded human judgments, automatic metrics, and the external Physics-IQ Verified benchmark.
\end{itemize}

\section{Related Work}
\paragraph{Self-improvement through generated data.} Learning from self-generated data offers a route to model adaptation with less human annotation~\citep{wang2023selfinstruct,zelikman2022star}. Two relevant directions are constructing training supervision from supplied knowledge or examples and autonomously generating new learning tasks~\citep{huang2026rzero}. For adaptation to supplied inputs, self-verification and feedback from downstream performance offer complementary ways to assess and improve the generated supervision~\citep{zeng2025kbalign,zweiger2025seal}. For autonomous task generation, an executable environment can validate proposed tasks and solutions, allowing task proposal and solving to be optimized together~\citep{zhao2025absolutezero}. For video models, VLM feedback guides offline fine-tuning to improve object interactions~\citep{furuta2024aifeedback}, while embodied video planning supports further training on successful trajectories collected from the environment~\citep{soni2024videoagent}. These approaches connect generated content to model updates through feedback on its validity or learning utility.

\paragraph{Physical supervision for video generation.} Controlled physical data connect known scene conditions with observed behavior, providing structured supervision for learning physical dynamics. Task-oriented datasets cover stability and support, temporal and causal reasoning, and physical prediction across object interactions~\citep{groth2018shapestacks,yi2020clevrer,bear2021physion}. A complementary direction varies latent physical properties to test whether models infer dynamics from observed motion and interactions~\citep{chen2022comphy,tung2023physionpp}. Reusable simulation platforms provide annotated scenes and multimodal observations, supporting data construction beyond individual tasks~\citep{greff2022kubric,gan2021threedworld}. A unified physical representation allows captured 3D Gaussian assets, meshes, and fluids to interact within the same simulation~\citep{liu2026raf}. Integrated suites combine physical domains and visual learning tasks~\citep{zhou2026physinone}, while simulation-based supervision and feedback from vision-language models support learning physical controls for video generation~\citep{narayanan2026phyco}.

\paragraph{LLM-guided simulation and physical video generation.} Work on language and vision-language agents spans interactive simulator access, programmatic scene construction, and coordinated generation of dynamic physical worlds~\citep{ren2025simworld,kang2026simworldstudio,yin2026viga,zhang2026gsagent}. Visual feedback and physical constraints help refine language-generated scene layouts while avoiding interpenetration and unstable support~\citep{ling2026scenethesis}. For video generation, simulated trajectories supply motion guidance~\citep{foo2026psivg}, while agentic planning uses tool execution and verification to revise generation decisions~\citep{feng2026newton}. Stage-specific verification of reconstructed scenes and simulated events supports targeted program repair before video synthesis~\citep{li2026physagent}. When simulated motion guides visual refinement, the final video still needs to preserve the dynamics verified in simulation.

\paragraph{Physical assessment and video correction.} Video assessment separates perceptual quality from physical behavior. Perceptual protocols evaluate appearance and temporal coherence~\citep{huang2024vbench}, while complementary benchmarks assess physical commonsense, rule adherence, and consistency with real-world principles~\citep{bansal2024videophy,bansal2026videophy2,meng2025phygenbench,zheng2025vbench2}. Reference-based tests compare generated motion with recorded experiments or simulated references~\citep{motamed2026physicsiq,radsch2026physicsiqverified,zhou2026physinone}, and temporally localized diagnoses identify when physical failures occur and explain them~\citep{zhang2026physioneval}. At inference, self-refining sampling revises denoising predictions without external verifiers~\citep{jang2026selfrefining}, while VLM feedback supports iterative refinement and selection of video plans for embodied tasks~\citep{soni2024videoagent}. Motion controllers instead guide generation through user-specified trajectories or optical flow~\citep{zhang2025motionpro,burgert2025gowiththeflow}. For existing videos, diagnosis-guided localized refinement corrects text--video misalignment while preserving faithful regions~\citep{lee2026videorepair}. Inpainting fills missing regions~\citep{zhou2023propainter,zhang2024avid}, and correspondence-based editing maintains consistency across frames~\citep{geyer2024tokenflow}. These methods provide different forms of control, but visual coherence or adherence to a prescribed motion alone does not establish physical correctness. PhysLoop connects diagnosis to simulation-grounded correction and evaluates both the resolution of the target physical error and the preservation of unaffected content.

\section{PhysLoop}
\label{sec:physloop}

\begin{figure*}[t]
\centering
\includegraphics[width=\linewidth]{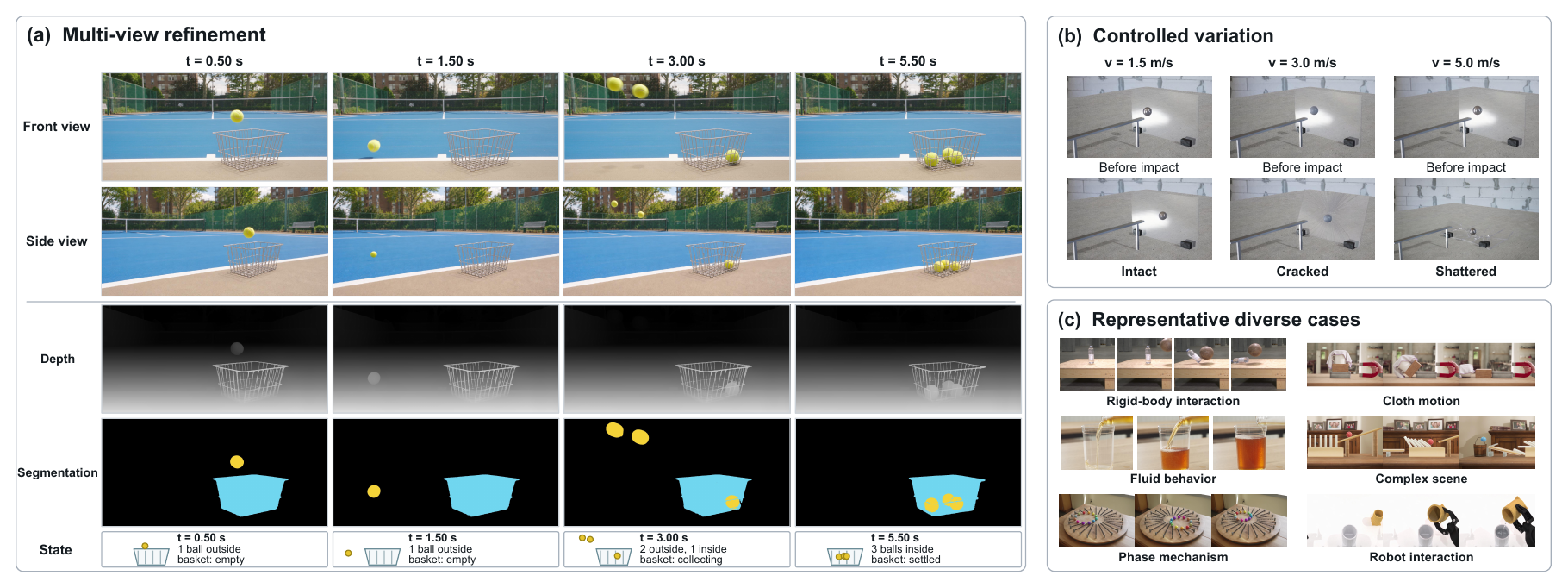}
\caption{Dataset overview: (a) multi-views, depth, segmentation, and states; (b) controlled variations; (c) representative diverse cases.}
\label{fig:physloop-data-gallery}
\end{figure*}

\subsection{Workflow}
\label{sec:physloop-overview}

PhysLoop takes a text request, optionally accompanied by image or video references, and returns a generated or repaired video together with its task specification and run record. As shown in Figure~\ref{fig:physloop-workflow}, the same LLM takes on six roles in the workflow: (1) writing the task specification, (2) acquiring assets and building the scene, (3) running simulation and rendering through the simulator, (4) verifying the simulation, (5) invoking a video generation model, the \emph{refiner}, to refine appearance, and (6) verifying the final video. Data construction and video repair share this workflow, with task constraints and outputs specific to each application (Sections~\ref{sec:physloop-data} and~\ref{sec:physloop-repair}).

\subsection{Task Specification and Scene Construction}
\label{sec:physloop-construction}

The agent expands the request into a task specification that records the target event, construction assumptions, and conditions used for verification. Text specifies the intended behavior, while images and videos supply visual references. For a falling-glass request, the agent must choose a release height and an impact surface when these are unspecified. These assumptions are recorded alongside the expected outcome so that construction and verification refer to the same task.

The agent retrieves assets that meet the visual and interaction requirements. When suitable assets are unavailable, it searches external sources or generates assets with Meshy~\cite{meshy_api}. The harness checks asset scale and collision geometry against the intended interaction and handles rendering meshes and collision proxies separately. The agent writes a scene program that combines retrieved assets and procedural geometry, specifying object placement and relationships. This allows the same event to be constructed in different settings and supports structures beyond the asset catalog. Before full simulation, the harness renders a low-resolution preview frame from each planned camera view. If an object is misplaced or an interaction is occluded, the agent adjusts the layout or camera and repeats the preview.

\subsection{Physics Execution and Video Refinement}
\label{sec:physloop-execution}

Unreal Engine (UE)~\cite{unreal_engine} is the primary simulator integrated into the harness, while Taichi~\cite{hu2019taichi} and Genesis~\cite{genesis2024} provide additional support. Given the initial conditions and permitted actions specified by the agent, the physics solver numerically computes the resulting motion according to the physical model of the scene. Simulator adapters are conversion interfaces that translate scene configurations into each simulator's inputs and transfer the resulting motion to UE with aligned coordinates and timing.

Multiple camera views share the same simulation timeline. Alongside RGB, the harness captures depth and instance segmentation and records object states and backend-provided contact or event signals. Each frame is linked to its simulation time and camera configuration, allowing verification to relate visible behavior to physical states.

After simulation checks pass, appearance refinement adapts the rendered scene to the requested visual appearance. The video refiner takes the verified simulation video as a motion reference, together with the task description and visual references that specify the desired appearance.

\subsection{Verification and Revision}
\label{sec:physloop-revision}

Successful execution does not establish that the requested event occurred. The verifier compares simulation states and visual observations with the predefined event requirements and physical conditions. For a collision task, it checks whether contact occurred and whether the subsequent motion satisfies the specification. Conservation checks account for external forces, dissipation, and numerical error.

The final-video check evaluates whether visible behavior preserves the target event and its physical relationships after refinement. Refinement may alter motion that passed simulation checks, so the verifier reassesses the refined result against the same event requirements.

If either simulation or final-video verification fails, the verifier returns a failure description with supporting simulation or visual evidence to the agent. Using this feedback and the recorded history, the agent revises the configuration for the responsible stage. A missed collision may require changes to the scene or initial conditions, while a refinement error may require different refiner inputs.

Execution ends when all required checks pass or a user-set budget on revision rounds and video-model calls is exhausted; if no candidate passes, PhysLoop returns the candidate with the fewest failed checks and marks its unresolved failures in the run record. The specification, configurations, observations, and revision history remain linked in this record. Human review is optional: users can review candidates during verification, and their judgments are recorded together with automated checks and agent revisions.

\subsection{Application: Training-Data Construction}
\label{sec:physloop-data}

The agent exposes task-specific parameters in the generated scene configuration, such as release height and initial velocity for a falling glass. The agent or user varies selected parameters while holding the remaining conditions fixed, producing different instances of the same event. Scene and camera settings can also be edited, and multiple views can be rendered from the same simulation.

For data construction, the task description and visual references specify the appearance to be produced. Videos that pass final verification are paired with their task descriptions as training examples for video generation models. Each sample retains its generating configuration for inspection, subsequent expansion, and regeneration. Downstream training can use these video--text pairs to update the video generation model.

Raw simulation renders retain their synchronized depth, segmentation, and object-state annotations. Because refinement may alter geometry or object positions, these annotations are not treated as pixel-aligned supervision for refined videos. The annotated simulation renders remain a separate resource from the refined video--text pairs.
\begin{table*}[t]
\centering
\small
\setlength{\tabcolsep}{4.2pt}
\renewcommand{\arraystretch}{1.10}

\begin{tabular*}{\textwidth}{
@{\extracolsep{\fill}}
l
cccc
cccc
ccc
@{}
}
\toprule
\multirow{2}{*}{Method}
& \multicolumn{4}{c}{Human (Gen)}
& \multicolumn{4}{c}{Model (Gen)}
& \multicolumn{3}{c}{Task} \\
\cmidrule(lr){2-5}
\cmidrule(lr){6-9}
\cmidrule(lr){10-12}

& Phys. & Real. & Causal & Win
& PC & SA & IQ & AQ
& Gen & Repair & P-IQ \\
\midrule

\multicolumn{12}{c}{\textit{Video Generation Models}} \\
\midrule

Gemini Omni
& 3.11 & 3.82 & 3.18 & 18.94
& 20.79 & \textbf{39.17} & 71.17 & 50.40
& 49.00 & 14.67 & 31.66 \\

H3
& 3.63 & 4.27 & 3.55 & 44.93
& \textbf{22.98} & 38.68 & 68.47 & 50.62
& 53.00 & 18.00 & 28.14 \\

Seedance 2.5
& 3.80 & \textbf{4.29} & 3.80 & 53.79
& 21.51 & 38.69 & 67.63 & \textbf{50.63}
& 73.00 & 19.33 & 48.07 \\

\midrule

\multicolumn{12}{c}{\textit{Seedance 2.5}} \\
\midrule

Claude Opus 5
& 3.82 & 3.82 & 3.84 & 64.18
& 20.23 & 36.69 & 67.66 & 49.71
& 71.00 & 78.67 & \textbf{53.75} \\

Kimi K3
& 3.63 & 3.69 & 3.69 & 54.41
& 19.90 & 37.32 & 68.54 & 50.42
& 69.00 & 68.67 & 52.04 \\

\midrule

\multicolumn{12}{c}{\textit{GPT-6 Astra}} \\
\midrule

Gemini Omni
& 3.53 & 2.96 & 3.51 & 45.45
& 12.71 & 35.84 & \textbf{75.11} & 46.72
& 67.00 & 62.00 & 12.08 \\

H3
& 3.86 & 3.45 & 3.90 & 58.21
& 17.48 & 37.12 & 68.77 & 48.65
& 73.00 & 72.67 & 39.10 \\

w/o Refiner
& 3.94 & 2.59 & 3.95 & 34.33
& 12.53 & 34.88 & 72.42 & 45.75
& 80.00 & 60.67 & -- \\

\midrule

\multicolumn{12}{c}{\textit{GPT-6 Astra + Seedance 2.5}} \\
\midrule

\textbf{PhysLoop}
& \textbf{4.06} & 3.72 & \textbf{4.12} & \textbf{75.76}
& 15.29 & 36.28 & 69.93 & 48.70
& \textbf{85.00} & \textbf{86.00} & 52.06 \\

\bottomrule
\end{tabular*}

\caption{Results on PhysLoop-Gen (Gen: task success, \%), PhysLoop-Repair (Repair: pass-vote rate, \%), and Physics-IQ Verified (P-IQ). Win: pairwise win rate over all nine configurations.}
\label{tab:physloop-main}
\end{table*}

\subsection{Application: Video Diagnosis and Repair}
\label{sec:physloop-repair}

For an existing generated video, the agent first locates the physical or event-level error and determines the event to reconstruct and the interval to replace. It inspects source frames around the error to identify the relevant objects, their spatial relationships, and their visible motion. The agent uses these observations and the intended correction to specify the scene layout, initial states, and target event. Any retained footage before or after the replacement interval constrains how the corrected segment connects to the source. During refinement, the simulation render provides a motion reference, and the source video provides an appearance reference for the reconstructed objects and environment.

Given a source video \(V\) with \(T\) frames, the repair replaces the selected half-open interval \([a,b)\) with the corrected segment \(\hat V_{a:b}\). The corrected segment matches the source resolution and frame rate and contains exactly \(b-a\) frames. The output is assembled as
\begin{equation}
V' = V_{0:a}\;\Vert\;\hat V_{a:b}\;\Vert\;V_{b:T},
\label{eq:physloop-assembly}
\end{equation}
where \(\Vert\) denotes concatenation at the original frame boundaries. The splice retains the source segments outside \([a,b)\). The same operation reconstructs the full clip when \(a=0\) and \(b=T\).

The verifier reviews the assembled video to determine whether the target error was corrected, whether the repaired segment connects coherently to any retained source footage, and whether the required source content was preserved. If refinement alters the intended physical behavior or a splice introduces a discontinuity, the agent uses the resulting evidence to guide further revision under the same stopping conditions.


\section{PhysLoop Dataset}
\label{sec:physloopdataset}

\subsection{PhysLoop-100K}
\label{sec:physloop-100k}

PhysLoop-100K covers broad physical categories, including kinematics, dynamics, rigid-body dynamics, magnetism, fluid mechanics, flexible and elastic materials, and fragmentation and fracture. For each case, the system automatically varies relevant physical parameters, such as mass and initial velocity, together with assets, scenes, lighting, and camera views. The dataset includes all valid combinations of the parameter settings defined for each case, allowing users to select configurations of interest.

PhysLoop-100K contains 100,000 accepted videos from 2,500 distinct physical events. Each event contributes an average of 40 videos from 3--5 viewpoints, with 2--4 key parameters varied across multiple values. Alongside refined video--text pairs and their generating configurations, the dataset provides the original simulation renders with their corresponding depth maps, instance segmentation maps, and object states. We split the dataset by source event, keeping all views and parameter variants of an event in the same split. Three annotators provided 7,500 judgments on a random sample of 2,500 accepted videos. Of these judgments, 6,225 (83.0\%) indicated both physical consistency with the task specification and preservation of the simulated motion after refinement. Treating individual judgments as independent, the 95\% Wilson confidence interval was 82.13\%--83.83\%. The supplementary material lists representative event families and examples of their parameter grids.

\subsection{PhysLoop-Gen}
\label{sec:physloop-gen}

PhysLoop-Gen addresses generation requests specifying a target physical event. Each instance contains a text request, with image references provided for some cases. The requests cover rigid-body, cloth, and fluid phenomena, ranging from simple bouncing or sliding to coupled interactions and multistage mechanical chains. Task categories and difficulty are defined from input properties, including interaction count, event ordering, and observable conditions, rather than from a particular model's performance.

PhysLoop-Gen contains 100 generation tasks. The tasks support comparisons among video generation approaches, including agentic frameworks such as PhysLoop and direct video generation models. Task descriptions and representative examples are provided in the supplementary material.

\subsection{PhysLoop-Repair}
\label{sec:physloop-repair-dataset}

PhysLoop-Repair addresses existing videos with identifiable physical or event-level errors. Each instance contains a source video with a physical error, an optional repair request, and annotations describing the error and its temporal interval. Annotations locate the error and specify conditions for a successful correction, such as maintaining a falling object's presence and motion consistent with the event conditions. The primary evaluation protocol supplies only the source video and request; error and interval annotations remain evaluation-only.

Source videos are generated by different video models, including Seedance series, MiniMax H3, and Gemini Omni 1.1 Flash, yielding 50 repair tasks. Inclusion requires an observable error, a specified correction, and source content to preserve. Evaluation checks target-error removal, connection to surrounding events, and preservation of unrelated content.

\begin{figure*}[th]
\centering
\includegraphics[width=\linewidth]{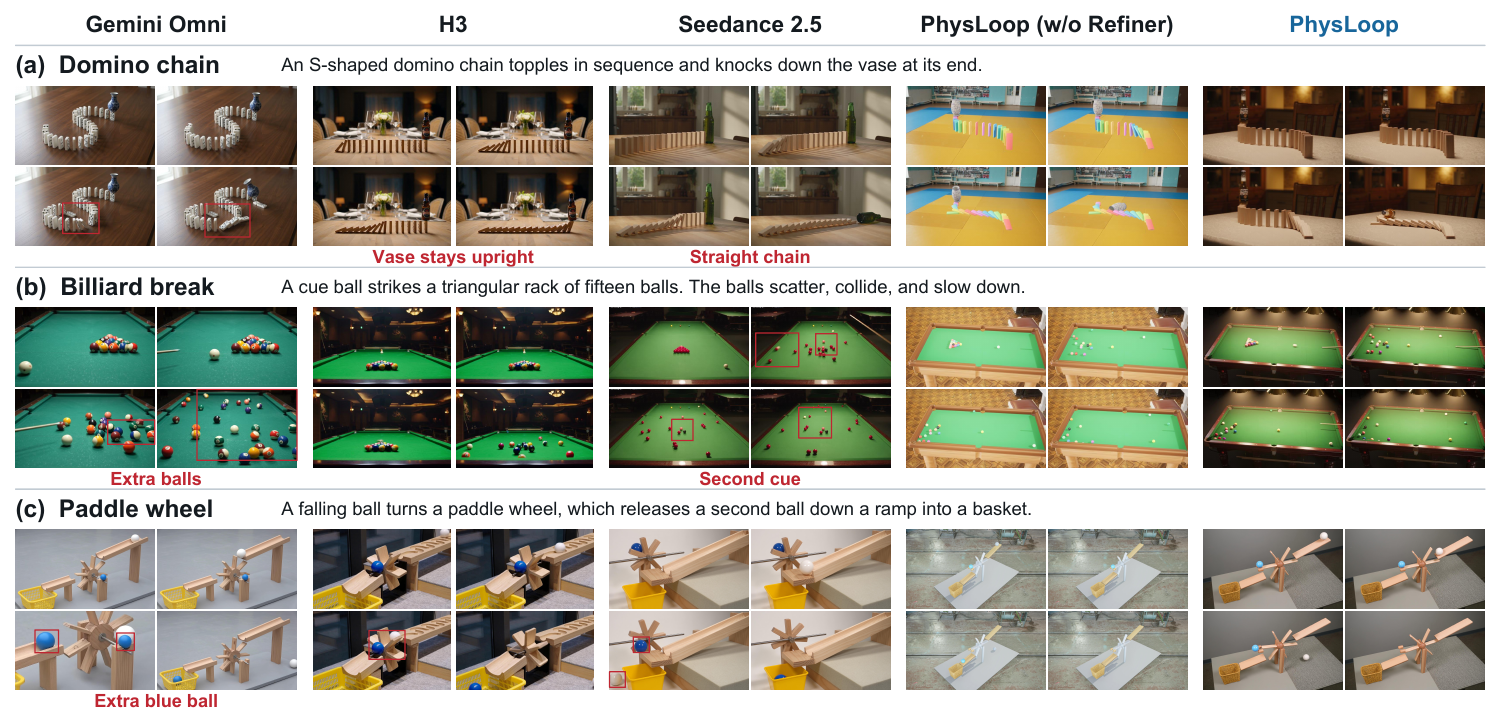}
\caption{Qualitative comparison of PhysLoop and baseline methods.}
\label{fig:physloop-evaluation-cases}
\end{figure*}



\section{Evaluation}
\label{sec:evaluation}

\subsection{Models and Experimental Settings}
\label{sec:evaluation-setup}

We evaluate video generation on PhysLoop-Gen, video repair on PhysLoop-Repair, and how well generated video continuations match recorded physical events on Physics-IQ Verified. The default configuration uses GPT-6 Astra~\citep{openai2026gpt6astra} as the agent backbone and Seedance 2.5~\citep{bytedance2026seedance25} for video refinement. With the refiner fixed, we compare the default backbone against Claude Opus 5~\citep{anthropic2026claudeopus5} and Kimi K3~\citep{moonshot2026kimik3}. With the backbone fixed, we compare the default refiner against MiniMax H3~\citep{minimax2026h3}, Gemini Omni (Gemini Omni 1.1 Flash)~\citep{google2026geminiomniflash}, and a no-refiner configuration.

Across the three benchmarks, each method may call its video generation model up to five times per task and returns one final output; PhysLoop configurations also allow at most ten simulation revisions, and no human review was used in these runs. For standalone baselines, we retain the first valid output. The supplementary material provides model versions and settings, prompts, output examples, costs, and runtimes.

\subsection{Metrics}
\label{sec:evaluation-metrics}

\paragraph{Human Evaluation.}

The five-point attribute ratings and automatic quality metrics apply to PhysLoop-Gen. PhysLoop-Repair is evaluated using the human pass-vote rate defined below. Human raters score physical plausibility, visual realism, and causal consistency on five-point scales. Overall win rate reflects pairwise preferences for physical and visual quality, with ties counting as half a win. Each video is assigned to multiple raters, with method identities hidden and presentation order randomized.

\paragraph{Automatic Evaluation.}
VideoPhy's VideoCon-Physics evaluator~\citep{bansal2024videophy} provides Auto-PC for physical commonsense and Auto-SA for adherence to the shared task description. VBench~\citep{huang2024vbench} measures imaging quality (IQ) and aesthetic quality (AQ). For each metric, we average scores over cases valid across compared configurations.

\paragraph{Task-Specific Metrics.}
For PhysLoop-Gen, a task passes if pass votes strictly outnumber fail votes, and the task success rate is computed over all 100 tasks. For PhysLoop-Repair, we divide pass votes by all valid pass and fail votes across the 50 tasks for each method.


Task-specific performance is the primary criterion, while human ratings provide the main assessment of generated-video quality. Compared with standalone Seedance 2.5, PhysLoop scores slightly lower on SA and AQ, lower on PC, and higher on IQ; the disagreement on physical plausibility may partly reflect limitations of automatic evaluation. Following prior work, we treat automatic metrics only as supporting references. They may favor visual styles common to video generators, while physical errors unfolding over time can be difficult to detect from sparsely sampled frames~\citep{bansal2026videophy2,zhou2026physinone}.

\subsection{Result Analysis}
\label{sec:evaluation-results}

\paragraph{Overall Comparison.}
Table~\ref{tab:physloop-main} summarizes results on PhysLoop-Gen, PhysLoop-Repair, and Physics-IQ Verified for standalone video generators and PhysLoop configurations. On PhysLoop-Gen, PhysLoop achieves an 85\% task-success rate, compared with 73\% for standalone Seedance 2.5. It also receives higher human ratings for physical plausibility (4.06 vs. 3.80) and causal consistency (4.12 vs. 3.80), although its visual realism rating is lower than that of standalone Seedance 2.5 (3.72 vs. 4.29). The gap in perceived realism is most apparent in material appearance and backgrounds.

On PhysLoop-Repair, the pass-vote rate was 86.00\% for PhysLoop and 19.33\% for standalone Seedance 2.5, the strongest standalone baseline. All three standalone video generators received pass votes on fewer than 20\% of valid judgments, showing that they struggle to satisfy the repair task’s combined requirements. A successful repair must correct the physical error while preserving the required source content and connecting coherently to surrounding events. PhysLoop expresses these requirements in an explicit reconstruction task and checks both the simulated event and the refined video against them. These results support the shared workflow of scene construction, simulation, and verification for correcting existing videos, extending its utility beyond controllable generation.
With GPT-6 Astra fixed, adding Seedance 2.5 refinement increased the generation task success rate from 80.00\% to 85.00\% and the repair pass-vote rate from 60.67\% to 86.00\%. Human-rated visual realism increases from 2.59 to 3.72, accompanied by slightly higher ratings for physical plausibility and causal consistency. Even without any video-model calls, the w/o Refiner configuration exceeds every standalone generator on PhysLoop-Repair (60.67\% vs.\ $\leq$19.33\%), so these gains do not come from repeated video sampling. Results also vary across refiners: H3 and Gemini Omni yield lower success rates on both tasks under the same backbone. The final video must satisfy both the requested appearance and the physical requirements of the task.

On Physics-IQ Verified, the default configuration scores 52.06, compared with 48.07 for standalone Seedance 2.5. With Seedance 2.5 fixed as the refiner, Claude Opus 5 achieves the highest score in our comparison (53.75), despite lower success on PhysLoop-Gen and PhysLoop-Repair. The differing rankings suggest that task completion and agreement with an observed physical continuation capture distinct aspects of performance. A plausible event may still differ from the specific dynamics of the reference video. Refiner choice also matters: with GPT-6 Astra fixed, H3 and Gemini Omni score 39.10 and 12.08, respectively, both below the default Seedance 2.5 configuration.

\subsection{Case Analysis}
\label{sec:evaluation-cases}


Figure~\ref{fig:physloop-evaluation-cases} shows why plausible local motion is insufficient for judging task completion. In the domino example, H3 depicts sequential toppling but leaves the vase upright, so the intended interaction chain remains incomplete. Standalone Seedance 2.5 replaces the requested S-shaped layout with a straight chain, violating the task’s spatial specification. In the billiard example, Gemini Omni introduces extra balls and Seedance 2.5 adds an extra cue ball, altering the object set required by the task. These failures concern the event outcome and scene consistency, which visual realism alone cannot establish. PhysLoop outputs with and without the refiner both depict the rack breaking apart, while differing in material appearance and lighting. This comparison illustrates how event realization and appearance refinement serve complementary purposes in the pipeline.

\section{Limitations}
\label{sec:limitations}

PhysLoop depends on the coverage of its simulators, assets, and verification conditions. Reconstructing a video requires assumptions about unobserved geometry, materials, and initial states, so a valid simulation may not recover the original physical process. Appearance refinement can introduce new errors, and local replacement may be insufficient when a mistake has already changed later events. These limitations constrain which videos can be accepted as learning signals and may require human review for ambiguous cases.

Extending PhysLoop to a complete RSI training loop requires connecting weakness discovery, candidate generation, verification and repair, data selection, and model updates across successive rounds. Each updated video model should influence which failures are targeted and which candidates are generated next. Future work must establish whether this process yields sustained gains on held-out events while preserving video quality and data diversity. Independent evaluation, tracking the effects of incorrect quality judgments, and controlling simulation and training costs will be necessary to support reliable self-improvement with decreasing human intervention.

\section{Conclusion}
\label{sec:conclusion}

We introduced PhysLoop, a harness that guides LLMs to realize their physical knowledge as videos under simulator constraints. One workflow constructs training data, detects physical errors, and repairs them by interval replacement, improving human-judged task success and repair over standalone generators at some cost in realism. These are the capabilities that recursive self-improvement requires before any model update: PhysLoop provides a quality-control component toward RSI, and fine-tuning video models on its data to close the loop is our next step.

\section*{LLM Usage Statement}
ChatGPT assisted in editing or refining author-produced writing, related work, code, experiments, figures, data, and research ideas. This assistance focused on refining existing material, with scientific decisions and interpretation remaining with the authors. GPT-6 Astra, Claude Opus 5, Kimi K3, Seedance, MiniMax H3, and Gemini Omni 1.1 Flash were used for experiments and data generation. The authors take full responsibility for the methodology, results, conclusions, and all submitted content.

\bibliographystyle{plainnat}
\setlength{\bibhang}{0pt}
\setlength{\bibindent}{0pt}
\setlength{\bibsep}{2pt}
\bibliography{references}

\appendix
\renewcommand{\thetable}{S\arabic{table}}
\renewcommand{\thefigure}{S\arabic{figure}}
\renewcommand{\theequation}{S\arabic{equation}}
\setcounter{table}{0}
\setcounter{figure}{0}
\setcounter{equation}{0}
\section{Data Resources and Construction}
\label{app:data}

\subsection{Training-Data Construction}
\label{app:training-data}

We construct training data by expanding physical event families into multiple scene configurations. For each family, we vary physical parameters such as mass, initial velocity, and material properties, together with object assets, backgrounds, lighting, and camera views. Each configuration undergoes simulation, verification, appearance refinement, and a final-video check before the output is paired with a text description. The training pair is linked to its generating configuration and native simulation records, including the rendered video, depth maps, instance segmentation, and object states.

Table~\ref{tab:app-physical-families} lists representative event families and the parameters varied to build their variants. A composite scene can combine several families; for example, a falling ball turns a wheel that releases a second ball into a basket.

\ifdefined\PhysLoopAppendixEmbedded
\begin{table*}[t]
\else
\begin{table}[htbp]
\fi
\centering
\small
\setlength{\tabcolsep}{5pt}
\renewcommand{\arraystretch}{1.15}
\begin{tabular}{@{}p{0.22\linewidth}p{0.35\linewidth}p{0.37\linewidth}@{}}
\toprule
Category & Representative events & Varied parameters \\
\midrule
Rigid-body motion and contact & Projectile motion, bouncing, sliding, rolling, billiard collisions & Mass, initial velocity, friction, restitution \\
Rigid-body mechanisms & Domino chains, pendulums, Newton's cradles & Release angle, constraint length, object spacing \\
Elastic and deformable systems & Elastic-body impact, cloth draping, flag motion & Stiffness, damping, Young's modulus, wind velocity \\
Magnetic interactions & Attraction and repulsion & Field mode, strength, initial separation \\
Fluids & Container transfer, spilling, swirling flow & Initial volume, release height, initial velocity field \\
Fracture & Glass-panel and wooden-crate impact & Impactor mass, launch velocity, impact position, damage threshold \\
\bottomrule
\end{tabular}
\caption{Representative physical event families and the parameters varied to construct their variants.}
\label{tab:app-physical-families}
\ifdefined\PhysLoopAppendixEmbedded
\end{table*}
\else
\end{table}
\fi

Each family defines a parameter grid in its scene configurations; Table~\ref{tab:app-parameter-grids} shows examples. Because a parameter change can shift both event timing and the final state, each variant keeps its own simulation timeline.

\ifdefined\PhysLoopAppendixEmbedded
\begin{table*}[t]
\else
\begin{table}[htbp]
\fi
\centering
\small
\setlength{\tabcolsep}{4pt}
\begin{tabularx}{\linewidth}{@{}lXX@{}}
\toprule
Family & Parameter & Values \\
\midrule
Ramp rolling/sliding & Friction (static, dynamic) & $0.02,\ 0.12,\ 0.55$ \\
Newton's cradle & Release angle & $25^\circ,\ 35^\circ,\ 45^\circ$ \\
Elastic-body impact & Young's modulus & $50,\ 100,\ 200$ kPa \\
Pinned flag in wind & Wind speed & $3.0,\ 6.5$ m/s \\
Fluid volume release & Initial center height & $0.55,\ 0.65,\ 0.75$ m \\
Glass-panel impact & Impactor speed & $1.5,\ 3.0,\ 5.0$ m/s \\
\bottomrule
\end{tabularx}
\caption{Example parameter grids from archived scene configurations.}
\label{tab:app-parameter-grids}
\ifdefined\PhysLoopAppendixEmbedded
\end{table*}
\else
\end{table}
\fi

Each sample record links the task, its scene configuration, the native observations, and the refined output. Native RGB, depth, and instance segmentation share timestamps and camera settings; object-state records contain identities, positions, orientations, and the contact and event records exposed by the simulator. Physics step rate and render frame rate are stored separately so that each saved frame maps to a simulation time. Depth and instance labels index the native render, not the refined video.

Multiple cameras observe the same simulation run. Cameras are placed around the objects and contacts that verification must observe, and a low-resolution layout preview checks scale, framing, and occlusion before the full capture. For example, the four-ball basket task is recorded from multiple viewpoints, with native RGB, depth, and instance segmentation.

\paragraph{Assets.}
Scenes combine assets from three sources: licensed content packages imported into the local catalog, Poly Haven environment maps and materials (CC0), and meshes generated with Meshy when no catalog asset fits. Every scene object records its asset source and license tier. Visual meshes are paired with separate collision proxies; for example, an open funnel uses a continuous conical collision shell matched to the visible inner wall.

\paragraph{Quality audit.} Each of the three annotators independently sampled and evaluated 2,500 accepted videos from PhysLoop-100K, with one video per physical event. The sampled videos could differ across annotators. Annotators saw the refined video, its task specification, and the native simulation render, and judged whether the refined video was physically consistent with the specification and whether it preserved the simulated motion. Of the 7,500 judgments, 6,225 (83.0\%) were positive on both criteria. Because every sampled video had passed the final verifier, this rate also measures how often the verifier's acceptance agrees with human judgment.

\subsection{PhysLoop-Gen}
\label{app:gen}

PhysLoop-Gen contains 100 tasks: 68 rigid-body tasks, 12 cloth tasks, 9 fluid tasks, and 11 image-conditioned tasks. The image-conditioned group is defined by its input modality and can involve rigid-body or other interactions. Each task contains a text request, optional image references, and success conditions that name the required objects, key interactions, event order, and final outcome.

Task difficulty is defined by the input: the number of interacting objects, required contacts, constraints, and event stages. Multistage tasks are checked stage by stage. In the image-conditioned paddle-wheel task (Fig.~\ref{fig:app-paddle-wheel-example}), a falling ball turns a wheel, whose blade pushes a second ball off its support and down a ramp into a basket.

\ifdefined\PhysLoopAppendixEmbedded
\begin{figure*}[t]
\else
\begin{figure}[htbp]
\fi
\centering
\includegraphics[width=\linewidth]{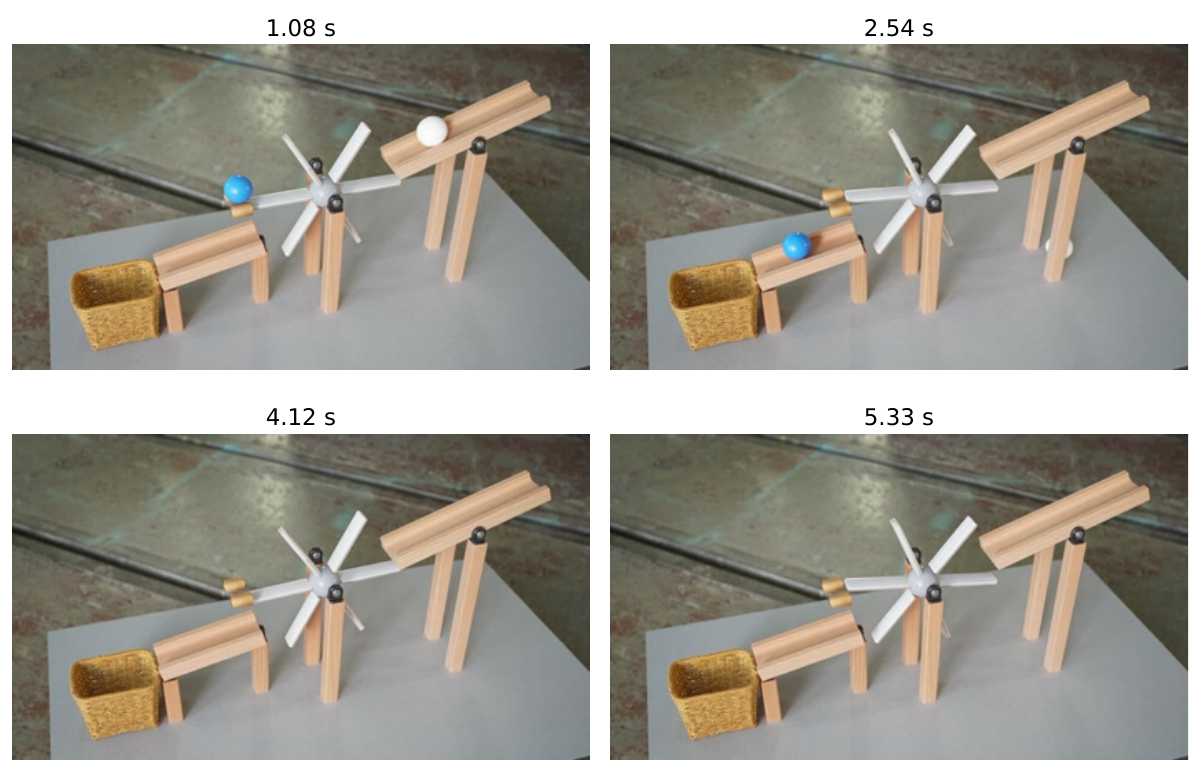}
\caption{Example frames of the paddle-wheel task.}
\label{fig:app-paddle-wheel-example}
\ifdefined\PhysLoopAppendixEmbedded
\end{figure*}
\else
\end{figure}
\fi

\subsection{PhysLoop-Repair}
\label{app:repair-data}

Each repair task contains a source video, an optional repair request, and evaluation-only annotations of the error, its interval, and the conditions for a successful correction. Error types include motion or contact failures, changes in object count or identity, missing support, wrong event order, and inconsistent material response.

Repairs follow the interval replacement defined in the main paper. A localized repair replaces the interval $[a,b)$ with exactly $b-a$ frames at the source resolution and frame rate, keeps blending inside the interval, and is checked for unchanged decoded frames outside it. When the error has already altered later events, the agent reconstructs the whole clip, which corresponds to $a=0$ and $b=T$. Each task records which scope was used.

A repair succeeds when the target error is removed, the result connects to the surrounding events, required source content is preserved, and no new severe error appears. Evaluation always inspects the complete output video.

\section{Harness Execution Details}
\label{app:harness}

\subsection{Verification Checks}
\label{app:verification}

Simulation-stage verification has three layers.

\textbf{Task assertions.} The task specification declares machine-checkable assertions over the native simulation record: existence, count, or order of contacts between named objects; thresholds on an object's state value or state change; trajectory integrity; and completeness of the required media and state artifacts. Contact evidence combines simulator hit events, when available, with oriented-box proximity tests using a 4\,cm tolerance. Proximity-based contacts are geometric inferences, not native solver contact impulses. For the paddle-wheel task, the assertions require contact between the driving ball and a wheel blade, contact between another blade and the second ball, a displacement of the second ball of at least 0.45\,m, and complete trajectories.

\textbf{Technical quality gate.} Independently of the task, the gate checks media validity (codec, frame count, frame rate), that each camera moves only as declared, that the initial scene is free of unintended contacts, that compiled object placements match the simulator's pre-simulation state, and, when a depth reference exists, that native depth agrees with the declared reference support surface (mean absolute error $\le$1\,cm, 95th percentile $\le$2\,cm). Joint frames must coincide within 2\,mm and $0.1^\circ$ at the initial pose.

\textbf{Semantic review.} The verifier, i.e., the agent LLM in its verification role, checks the evidence bundle against the task conditions. The bundle contains event-selected keyframes from each camera, the contact records, sampled trajectories and constraint states, and the quality-gate results. After refinement, the final-video check applies the same task conditions to the refined video, because refinement can alter motion that passed simulation.

\subsection{Budgets, Stopping, and Human Involvement}
\label{app:budgets}

Each task runs under a user-configurable controller budget that limits scene-specification revisions, simulator launches, technical retries, and wall-clock time. Configured limits and consumed usage are recorded for each run. A run stops when all checks pass or a limit is reached; if no candidate passes, the candidate with the fewest failed checks is returned, and unresolved failures stay in the run record. In the main-paper experiments, each task allows at most ten scene-specification (simulation) revisions, and both PhysLoop and the standalone baselines permit at most five video-generation calls per task.

Users can review candidates during verification. Each human judgment, the change it requested, and the resulting revision are stored in the run record, so automated and human-requested revisions can be counted separately.

\section{Experimental Settings and Evaluation Protocols}
\label{app:evaluation}

\subsection{Compared Configurations}
\label{app:configurations}

Table~\ref{tab:app-configurations} lists the nine configurations. Each provides one final output for each of the 100 PhysLoop-Gen tasks, for 900 videos in total.

\ifdefined\PhysLoopAppendixEmbedded
\begin{table*}[t]
\else
\begin{table}[htbp]
\fi
\centering
\small
\setlength{\tabcolsep}{4pt}
\begin{tabularx}{\linewidth}{@{}XX@{}}
\toprule
Agent & Video model \\
\midrule
-- (text only) & Gemini Omni 1.1 Flash \\
-- (text only) & Seedance 2.5 \\
-- (text only) & MiniMax H3 \\
Claude Opus 5 & Seedance 2.5 \\
Kimi K3 & Seedance 2.5 \\
GPT-6 Astra & Gemini Omni 1.1 Flash \\
GPT-6 Astra & Seedance 2.5 \\
GPT-6 Astra & MiniMax H3 \\
GPT-6 Astra & -- (w/o Refiner; native render) \\
\bottomrule
\end{tabularx}
\caption{Compared configurations. Each row contributes 100 final outputs.}
\label{tab:app-configurations}
\ifdefined\PhysLoopAppendixEmbedded
\end{table*}
\else
\end{table}
\fi

The experiments use GPT-6 Astra, Claude Opus 5, and Kimi K3 as agent backbones. All 900 final outputs in PhysLoop-Gen are encoded at 24\,fps and range from 5.00 to 8.04\,s in duration. Gemini Omni 1.1 Flash and Seedance 2.5 outputs have a resolution of \(1920\times1080\), and MiniMax H3 outputs have a resolution of \(1344\times768\). The w/o Refiner configuration uses native UE renders outputs have a resolution of \(1920\times1080\).

\subsection{Cost and Runtime Accounting}
\label{app:cost}

We recorded token usage on ten PhysLoop-Gen tasks stratified by task category and number of event stages, using GPT-6 Astra and Seedance 2.5. Each task ran in a separate session, and the usage totals include failed attempts, revisions, and rejected requests. We ran simulation and rendering in UE~5.7 on a Mac mini with an Apple M4 chip and 16\,GB of memory, reusing cached assets. Seedance produced 1080p videos at 24\,fps.

Figure~\ref{fig:app-cost-runtime}(a) compares costs by applying the short-context rates of GPT-5.6 Sol and GPT-6 Astra to the same recorded token usage; the GPT-5.6 Sol column is hypothetical, because the runs used GPT-6 Astra and another model would consume a different number of tokens. The ten tasks used 199.05 million input tokens, including 196.53 million cached input tokens, and 0.58 million output tokens, including reasoning tokens. The rates per million uncached input, cached input, and output tokens are \$4, \$0.40, and \$20 for GPT-5.6 Sol and \$10, \$1, and \$50 for GPT-6 Astra.\footnote{\url{https://developers.openai.com/api/docs/pricing}.} The estimated agent costs are \$10.03 and \$25.07 per final video, respectively. Adding the same \$4.30 per video for generation and upload gives estimated totals of \$14.33 and \$29.37, respectively.

Figure~\ref{fig:app-cost-runtime}(b) shows runtime by stage. UE simulation and rendering took approximately 2--6 minutes, with a median of 3\,min 25\,s across 20 records. The low-resolution pre-check took about 3 minutes. Seedance refinement took approximately 2--3 minutes, with a median of 2\,min 13\,s across 28 records. Adding the three stage estimates gives a total of 7--12 minutes, excluding scene construction.

\ifdefined\PhysLoopAppendixEmbedded
\begin{figure*}[t]
\else
\begin{figure}[htbp]
\fi
\centering
\includegraphics[width=\linewidth]{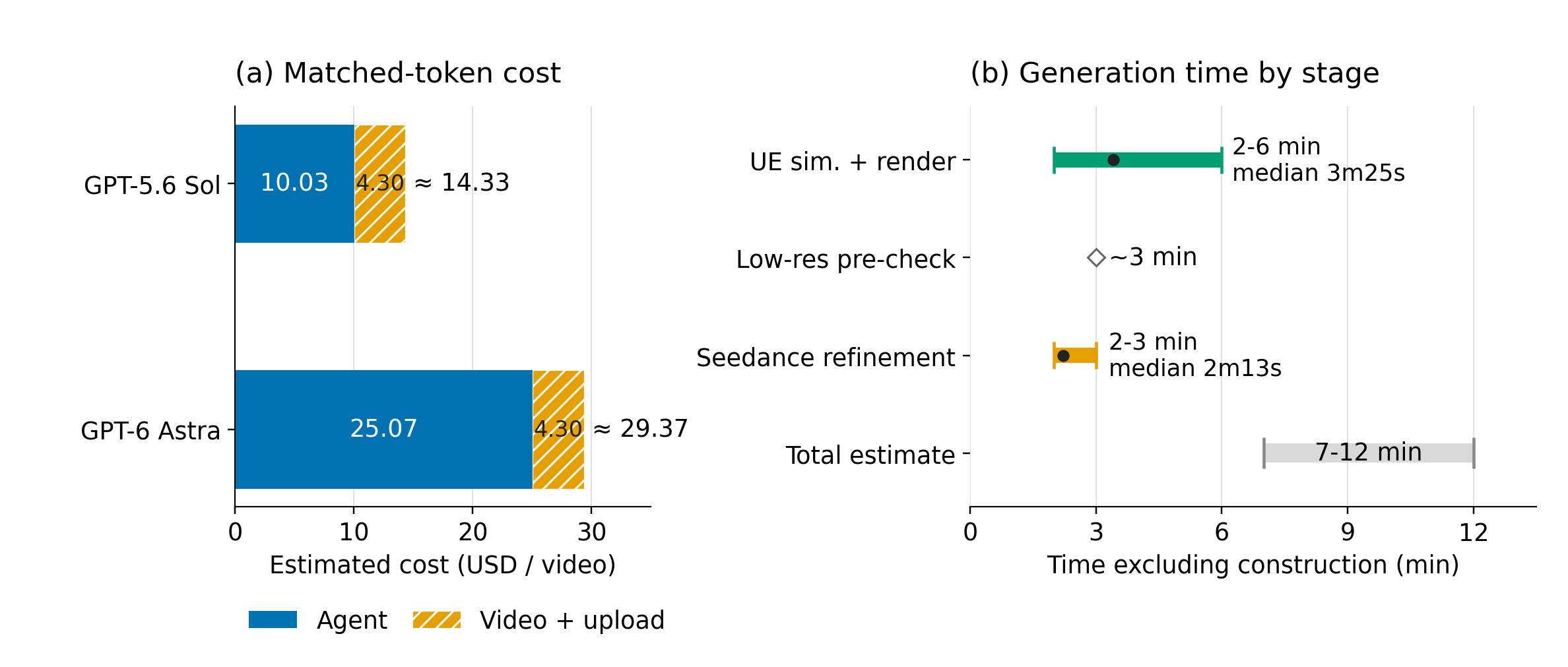}
\caption{Matched-token cost estimates and generation times excluding scene construction.}
\label{fig:app-cost-runtime}
\ifdefined\PhysLoopAppendixEmbedded
\end{figure*}
\else
\end{figure}
\fi

\subsection{Automatic Evaluation}
\label{app:automatic}

Physical commonsense (PC) and semantic adherence (SA) are scored by VideoCon-Physics on 32 sampled frames together with the task description. Imaging quality (IQ, MUSIQ-SPAQ) and aesthetic quality (AQ, LAION predictor) are computed by VBench on all frames. Stored scores are normalized to $[0,1]$ (MUSIQ divided by 100, aesthetic score divided by 10) and multiplied by 100 for display.

For configuration $m$ and metric $q$, scores are averaged over the cases $\mathcal{C}_q$ that are valid in every configuration:
\begin{equation}
\bar{s}_{m,q}=\frac{100}{|\mathcal{C}_q|}\sum_{c\in\mathcal{C}_q}s_{m,c,q}.
\label{eq:app-common-score}
\end{equation}

\subsection{Human Evaluation}
\label{app:human}

Raters see the task description and one video, then give separate integer scores from 1 to 5 for physical plausibility, visual realism, and causal consistency. Figures~\ref{fig:app-single-interface} and~\ref{fig:app-pair-interface} show the annotation interfaces. Raters judge observable evidence at normal playback speed: attractive appearance should not raise the physical-plausibility score, and physical correctness should not be equated with visual realism. N/A is used when an attribute cannot be judged; it is not a zero score.

Task completion is judged independently as pass, fail, or unjudgeable. An explanation is required for an unjudgeable response; raters can also explain their other judgments. A pass requires completion of the prompt's core event with plausible key physical relationships and causal order. Unjudgeable responses are excluded from the valid-vote denominator. Pairwise items show two outputs for the same task and ask for an overall preference (A, B, tie, or unjudgeable), considering all three attributes and task fulfillment without mechanically adding the scores. Method identities are hidden and presentation order is randomized.

Attribute scores are averaged first over the valid ratings of each video and then over videos, so each video has equal weight; N/A responses are treated as missing:
\begin{equation}
H_{m,a}=\frac{1}{|\mathcal{O}_{m,a}|}\sum_{c\in\mathcal{O}_{m,a}}\frac{1}{|\mathcal{R}_{m,c,a}|}\sum_{r\in\mathcal{R}_{m,c,a}}h_{m,c,r,a},
\label{eq:app-human-mean}
\end{equation}
where $\mathcal{R}_{m,c,a}$ are the valid ratings of attribute $a$ for case $c$ under configuration $m$, and $\mathcal{O}_{m,a}$ are the cases with at least one valid rating. Pairwise preference counts a tie as half a win and excludes unjudgeable items,
\begin{equation}
P_m=\frac{W_m+0.5\,T_m}{W_m+T_m+L_m}.
\label{eq:app-pairwise}
\end{equation}

For PhysLoop-Gen, we report task success rate as the number of passed tasks divided by 100. A task passes if pass votes strictly outnumber fail votes after excluding unjudgeable votes. For PhysLoop-Repair, we report the pass-vote rate as the number of pass votes divided by the total number of valid votes across the 50 tasks for each method.

\ifdefined\PhysLoopAppendixEmbedded
\begin{figure*}[t]
\else
\begin{figure}[htbp]
\fi
\centering
\includegraphics[width=\linewidth]{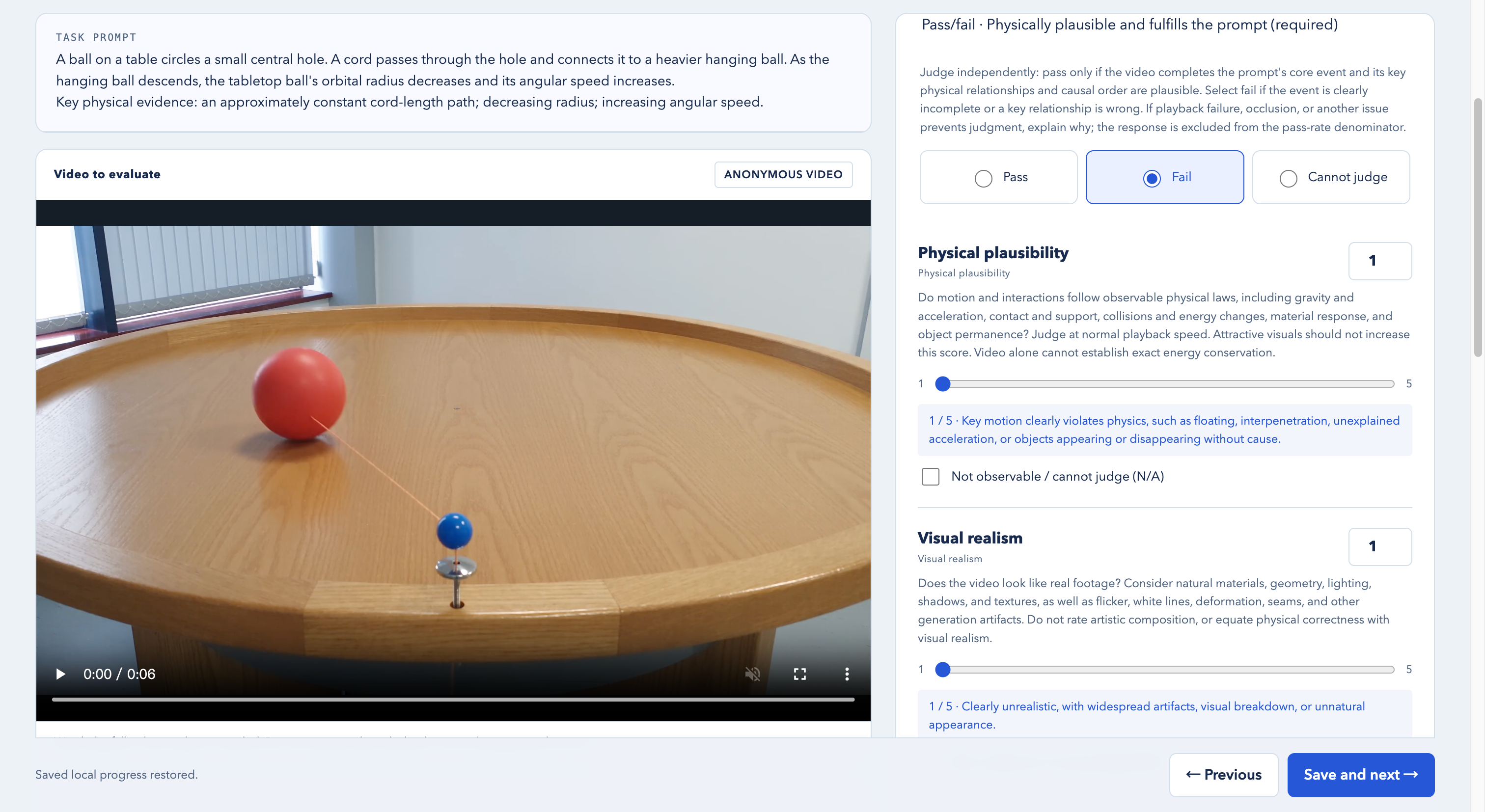}
\caption{Single-video evaluation interface for task completion and attribute ratings.}
\label{fig:app-single-interface}
\ifdefined\PhysLoopAppendixEmbedded
\end{figure*}
\else
\end{figure}
\fi

\ifdefined\PhysLoopAppendixEmbedded
\begin{figure*}[t]
\else
\begin{figure}[!htbp]
\fi
\centering
\includegraphics[width=1\linewidth]{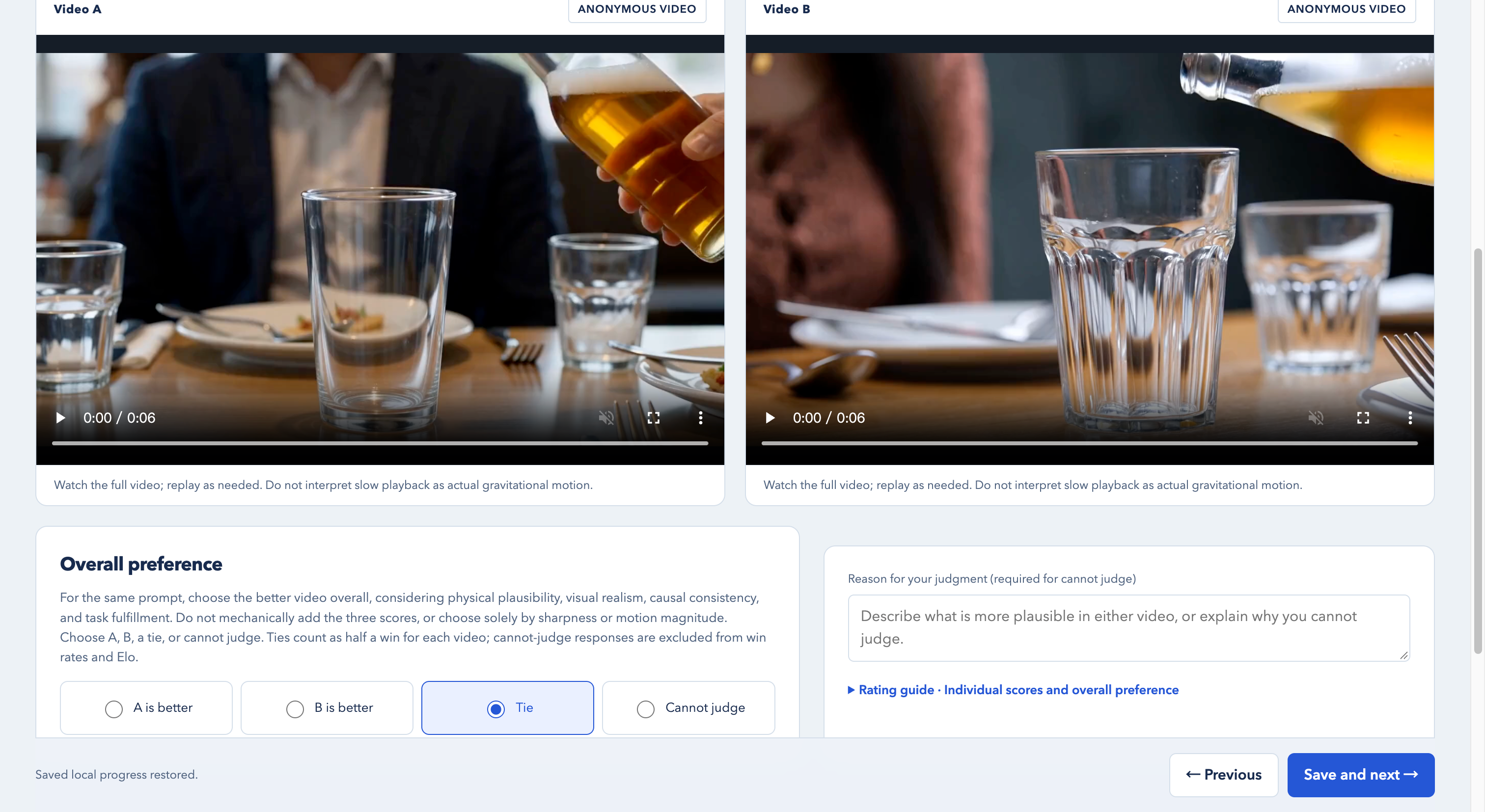}
\caption{Pairwise evaluation interface with A, B, tie, and cannot-judge options.}
\label{fig:app-pair-interface}
\ifdefined\PhysLoopAppendixEmbedded
\end{figure*}
\else
\end{figure}
\fi

\ifdefined\PhysLoopAppendixEmbedded\else\FloatBarrier\fi
\subsection{Physics-IQ Verified}
\label{app:physics-iq}

On Physics-IQ Verified, we use the first 120 output frames at 24\,fps for each continuation. Direct continuation methods receive a three-second recorded context clip, while UE-based methods reconstruct the scene from the transition frame and a scene description. The scores measure agreement between generated and recorded continuations under each method's input conditions.

\section{Representative Prompts and Skill Rules}
\label{app:prompts-skills}

We use the paddle-wheel task (Fig.~\ref{fig:app-paddle-wheel-example}) to illustrate the prompts at each stage.

\paragraph{Task request.}
\begin{quote}
A steel ball falls from a short chute onto one blade of a six-bladed rigid paddle wheel. The wheel rotates about a horizontal axle, and another blade knocks a colored ball off its support. The second ball rolls down a short ramp into a basket.
\end{quote}

\paragraph{Scene-construction request.}
\begin{quote}
Use the reference image for the layout of the wheel axle, blades, both balls, the support, the ramp, and the basket. The first ball must hit the wheel before it rotates; a blade must then hit the second ball, which leaves its support and enters the basket. The video must be at least five seconds long.
\end{quote}

\paragraph{Refinement prompt (Seedance 2.5).}
\begin{quote}
\emph{Scene and assets:} a wooden paddle-wheel chain in a real science museum: the original wooden chute, a six-bladed wheel on a horizontal axle, a white and a blue rubber ball, a short wooden ramp, and a yellow basket. Turn the static supports in the reference into sturdy wooden stands and metal bearing blocks; every raised part must have a visible load-bearing connection. Do not add moving parts or block the ball paths.

\emph{Motion:} the white ball rolls down the upper-right chute and hits the wheel, which turns about its axle; a blade knocks the blue ball down the lower-left ramp into the basket; the white ball falls to the right onto the display base and rolls out of frame. Motion and contact timing follow the reference exactly.

\emph{Shared constraints:} use the reference video as the only source of motion. Keep the number, identity, initial placement, trajectories, speed changes, contact timing, and final state of all objects. Change only the scene, appearance, and materials described above; do not add, duplicate, merge, or remove objects, and do not reorder or repeat events. Keep the reference camera and framing, with no text, cuts, or music. After the reference becomes still, hold the final state until the end.
\end{quote}

\paragraph{Verification assertions.}
\begin{quote}
\texttt{event\_exists}: contact(\texttt{drive\_ball}, \texttt{blade\_3});
\texttt{event\_exists}: contact(\texttt{blade\_1}, \texttt{second\_ball});
\texttt{state\_delta}: $\Delta x_{\mathrm{second\_ball}} \ge 0.45$\,m;
\texttt{trajectory\_integrity}; \texttt{artifact\_complete}.
\end{quote}

\paragraph{Skill rules.}
The agent follows a condensed \texttt{run-physics-video} skill:
\begin{enumerate}
\setlength{\itemsep}{0pt}
\setlength{\topsep}{2pt}
\setlength{\parskip}{0pt}
\setlength{\parsep}{0pt}
\item Before continuing a task, \texttt{inspect} its state, stage, blocker, and allowed actions; keep the original request and references unchanged across revisions.
\item Submit an intent and scene specification that match the current generation context; the controller validates them and records execution.
\item Check placement and framing with \texttt{layout-preview}, dynamics with solver-backed smoke runs, and later-time framing with \texttt{replay-preview} on the cached trajectory.
\item Revise only the configuration fields implicated by the failure evidence; task conditions, acceptance rules, and usage counters are preserved. Technical retries resume from the last checkpoint.
\item Invoke \texttt{review} separately once the evidence bundle exists, and report completion only after the controller reaches \texttt{completed}.
\end{enumerate}

\end{document}